\documentclass[review]{elsarticle}

\usepackage{stmaryrd}
\usepackage{bussproofs}
\usepackage{multirow}

\usepackage{llncsdoc}

\usepackage{mathptmx}

\usepackage{latexsym}
\usepackage{amsfonts}
\usepackage{amssymb}
\usepackage{amsmath}
\usepackage{graphicx}
\usepackage{color}

\usepackage{wrapfig,lipsum,booktabs}
\usepackage{semantic}
\usepackage{listings}
\usepackage{proof}

\usepackage{subcaption}
\usepackage{url}

\usepackage{tikz}
\usepackage{mathdots}
\usepackage{yhmath}
\usepackage{cancel}
\usepackage{siunitx}
\usepackage{array}
\usepackage{gensymb}
\usepackage{tabularx}
\usepackage{booktabs}
\usetikzlibrary{fadings}

\usepackage{pifont}
\usepackage{floatrow}
\usepackage{subcaption}

\newfloatcommand{capbtabbox}{table}[][\FBwidth]

\makeatletter
\newcommand\tabcaption{\def\@captype{table}\caption}
\newcommand\figcaption{\def\@captype{figure}\caption}
\makeatother

\newcommand{\N}{\mathbb{N}}

\newcommand{\cmark}{\ding{51}}
\newcommand{\xmark}{\ding{55}}

\definecolor{codegreen}{rgb}{0,0.6,0}
\definecolor{codegray}{rgb}{0.5,0.5,0.5}
\definecolor{codepurple}{rgb}{0.58,0,0.82}
\definecolor{backcolour}{rgb}{0.95,0.95,0.92}

\lstdefinestyle{csp}{
    backgroundcolor=\color{backcolour},   
    commentstyle=\color{codegreen},
    keywordstyle=\color{magenta},
    numberstyle=\tiny\color{codegray},
    stringstyle=\color{codepurple},
    basicstyle=\footnotesize,
    breakatwhitespace=false,         
    breaklines=true,                 
    captionpos=b,                    
    keepspaces=true,                 
    numbers=left,                    
    numbersep=5pt,                  
    showspaces=false,                
    showstringspaces=false,
    showtabs=false,                  
    tabsize=4,
    morekeywords={var,define,Stop,Skip,if,else,assert}
}

\DeclareMathOperator*{\minimise}{minimise}

\newcolumntype{L}[1]{>{\raggedright\arraybackslash}p{#1}}
\newcolumntype{C}[1]{>{\centering\arraybackslash}p{#1}}
\newcolumntype{R}[1]{>{\raggedleft\arraybackslash}p{#1}}

\newcommand{\limp}[0]{\rightarrow}

\newtheorem{thm}{Theorem}

\newdefinition{rmk}{Remark}
\newproof{pf}{Proof}
\newproof{pot}{Proof of Theorem \ref{thm2}}

\begin{document}


\title{Silas: High Performance, Explainable and Verifiable Machine Learning} 

%
%
%
%


\begin{frontmatter}

\author{Hadrien Bride, Zh\'e H\'ou}
\address{Griffith University, Nathan, Brisbane, Australia}

\author{Jie Dong}
\address{Dependable Intelligence Pty Ltd, Brisbane, Australia}

\author{Jin Song Dong}
\address{National University of Singapore, Singapore}

\author{Ali Mirjalili}
\address{Griffith University, Nathan, Brisbane, Australia}

\begin{abstract} 

This paper introduces a new classification tool named Silas, which is built to provide a more transparent and dependable data analytics service. A focus of Silas is on providing a formal foundation of decision trees in order to support logical analysis and verification of learned prediction models. This paper describes the distinct features of Silas: The Model Audit module formally verifies the prediction model against user specifications, the Enforcement Learning module trains prediction models that are guaranteed correct, the Model Insight and Prediction Insight modules reason about the prediction model and explain the decision-making of predictions. We also discuss implementation details ranging from programming paradigm to memory management that help achieve high-performance computation. 


\end{abstract}

\end{frontmatter}

\section{Introduction}
\label{sec:intro}

Machine learning has enjoyed great success in many research areas and industries, including entertainment~\cite{Gomez-Uribe2015}, self-driving cars~\cite{Eliot2017}, banking~\cite{Turkson2016}, medical diagnosis~\cite{Kononenko2001}, shopping~\cite{cumby2004}, and among many others. However, the wide adoption of machine learning raises the concern that most people use it as a ``black-box'' in their data analytics pipeline. The ramifications of the black-box approach are multifold. First, it may lead to unexpected results that are only observable after the deployment of the algorithm. For instance, Amazon's Alexa offered porn to a child~\cite{nypost2016}, a self-driving car had a deadly accident~\cite{guardian2018}, etc. Some of these accidents result in lawsuits or even lost lives, the cost of which is immeasurable. Second, it prevents the adoption in some applications and industries where an explanation is mandatory or certain specifications must be satisfied. For example, in some countries, it is required by law to give a reason why a loan application is rejected.


In recent years, eXplainable AI (XAI) has been gaining attention, and there is a surge of interest in studying how prediction models work and how to provide formal guarantees for the models. A common theme in this space is to use statistical methods to analyse prediction models. On the other hand, Bonacina recently envisaged that \emph{automated reasoning} could be the key to the advances of XAI and machine learning~\cite{bonacina2017automated}. This aligns well with our interest of building a new machine learning tool with logic and reasoning as the engine to produce ``white-box'' prediction models. A ``white-box'' machine learning method in our vision should feature the following key points:

\begin{description}
    \item[Explainability]: The inner workings of produced predictive models should be interpretable and users should be able to query the rationale behind their predictions.
    
    \item[Verifiability]: The validity of the produced predictive models with respect to user specifications should be formally verifiable.
    
    \item[Interactability]: Data engineers should be able to guide the learning phase of prediction models so that they conform with given specifications.
\end{description}

Towards this direction, we have been searching for a suitable machine learning technique that (1) has good predictive performance and (2) is suitable for logical reasoning and formal verification. We have found that some techniques show excellent performance but are difficult to understand, such as neural networks. Some techniques are easy to explain, e.g., linear methods, but often do not perform as well as the state-of-the-art. Some techniques have a solid probabilistic reasoning foundation, e.g., Bayesian methods, but are not suitable to reason about using formal logic. Some techniques that are interactive by nature, e.g., reinforcement learning, but are not in the supervised learning scope of this work. Between neural networks and ensemble trees, we choose the latter mainly because it is more amiable to logical reasoning, it has excellent predictive performance, sometimes better than deep learning~\cite{szilardbenchmark}, on \emph{tabular data}, and it requires less data-preprocessing. 

The current gap in the literature is the lack of understanding of the internal mechanism of ensemble trees and their \emph{perceived} black-box nature, which make them impractical in critical applications (e.g. medicine, law, Defence etc.) as discussed above. This gap motivated our attempts to reinvent a new classification tool named \emph{Silas}, which is a fusion of ensemble trees machine learning and automated reasoning and is built to provide transparent data analytics. We aim to make Silas a high-performance alternative to existing machine learning tools for supervised learning of structured data with an emphasis on dependability and transparency. Silas has the following novelties:

    First, Silas builds decision trees that are defined in a logical language that is designed for formal reasoning and verification. This theoretical foundation enables it to produce prediction models for which automated reasoning techniques can be leveraged to provide the following features: 

\begin{description}   
    \item[Model Audit]: Formally verify the correctness and safety of \emph{very large} prediction models in order to provide strong guarantees.

    \item[Enforcement Learning]: Train prediction models that are correct-by-construction with respect to user specifications.

    \item[Model Insight]: Analyse the prediction model and give a general idea of how the model makes predictions on each class.

    \item[Prediction Insight]: Explain the decision-making of individual predictions by relating them to their significant predictors.
    



\end{description}

    Second, Silas targets high-performance applications. Its predictive performance is on par with industrial leaders of similar techniques. Moreover, its C++ implementation is efficient and outperforms competitors in term of time and memory consumptions. As machine learning becomes more and more prevalent in everyday applications, Silas will provide increased productivity with lower operating costs.

The remainder of the paper is organised as follows: Section~\ref{sec:related} discusses the literature and how they relate to Silas. Section~\ref{sec:ml} describes the machine learning foundation of Silas. 
Section~\ref{sec:reason} discusses the white-box aspects of Silas with case studies using public datasets. 
Section~\ref{sec:comp} gives experimental results on Silas performance for large datasets. It also details implementation choices and illutrate their impact on computational performance. 
Finlay, section~\ref{sec:conc} concludes the paper.

\section{Related Work}
\label{sec:related}

There are many implementations of ensemble trees, such as xgBoost~\cite{Chen2016}, H2O~\cite{h2o} and Ranger~\cite{Wright2015}. The latter two are more relevant to the bagging implementation of Silas. H2O is a Java implementation that is shown more efficient than other tools~\cite{szilardbenchmark}, and it supports distributed computing. Ranger is a fast implementation of random forest written in C++ that is designed to handle high dimensional data. It is non-trivial to introspect and extract logical semantics from the structure of decision trees in popular tools. Thus, we have developed our own implementation of ensemble trees~\cite{Bride18,depintel} by using a tree structure that is amiable to logical reasoning. We show that our implementation is much faster and more memory efficient than both H2O and Ranger. The literature on ensemble trees and machine learning is rich and we will only focus on a subset that is related to the interpretability and verification of machine learning.

Although not yet substantial, there have been early steps taken towards understanding prediction models and providing guarantees for them. For instance, the Lime tool~\cite{lime2016} is able to provide local linear approximations of various types of prediction models and show which features are the most decisive in predictions. Similarly, Hara and Hayashi~\cite{Hara2016} proposed post-processing for ensemble trees to obtain an approximation of the model with probabilistic interpretations. Another interesting work is Lundberg et al.'s SHAP method~\cite{Lundberg2017}, which uses the game theory to obtain \emph{consistent} explanations. Ehlers~\cite{Ehlers2017} developed an SMT based method to verify linear approximations of feed-forward neural networks. While these methods have shown potential in interpreting and verifying predictions, they still treat the prediction model as a black-box and try to analyse or verify an approximation of the black-box. On the contrary, we are interested in treating the prediction model as a white-box and studying the internal mechanism of prediction models. 

A logical approach seems more natural for understanding the internal structure of decision trees because decision trees are inherently connected with logical semantics and are very similar to binary decision diagrams (BDDs) which are widely-used in implementations of logical systems such as theorem provers~\cite{Gore2014} and model checkers~\cite{Cimatti2002}. Caruana et al.'s work~\cite{Caruana2015} attempts to explain how a boosting machine makes predictions by analysing the logical conditions in the decision trees. However, at the time of writing their new Microsoft project was only 2 months old and the cited paper did not give enough details on interpretability.

Complementary to the above work, we are also interested in providing formal guarantees for prediction models. T\"ornblom and Nadjm-Tehrani~\cite{Tornblom2019} proposed a method to extract equivalent classes from random forest and verify that the input/output of the model satisfies safety properties. Their approach considers all possible combinations of results from all the trees, which means they have to verify $2^{d\cdot B}$ equivalent classes of the results where $d$ is the depth of trees and $B$ is the number of trees. The advantage of their approach is that they can give bi-directional results: (completeness) if the constraint is satisfied, their verification returns positive, and (soundness) if the verification returns positive, the constraints must be satisfied. The disadvantage of their approach is the high complexity and the verification of 25 trees of depth 20 in practice. Our verification approach focuses on soundness, as a result, we can simplify and parallelise the verification in order to verify very large models.  




\begin{table}[t!]
\begin{tabular}{lllll}
\hline
Methods/Tools   & Prediction & Explanation    & Verification & \begin{tabular}[c]{@{}l@{}}Correct-By-\\ Construction\end{tabular} \\ \hline
Neural Networks & \cmark        & \cmark*      & \xmark           & \xmark                                                                 \\
Ensemble Trees  & \cmark        & \cmark* & \cmark$\dagger$           & \xmark                                                                 \\
Silas           & \cmark        & \cmark            & \cmark          & \cmark                                                                \\ \hline
\end{tabular}
\caption{A comparison of key features of interest on some machine learning methods and tools. Correct-By-Construction means the ability to train models that are guaranteed correct w.r.t. user specifications. *For neural networks and other ensemble trees implementations, the explanation feature can be achieved via additional packages such as LIME and SHAP. $\dagger$Current verification methods for ensemble trees is not feasible to verify large models.}
\label{tab:related-comp}
\end{table}

Table~\ref{tab:related-comp} gives a comparison of key features of interest for this paper in popular machine learning methods and tools. Note that we only consider the verification of the full prediction model instead of the verification of an approximation of the model. In this work, we focus on providing a white-box analysis for \emph{binary classification} as the stepping stone for more general data analysis tasks. The remainder of the paper describes technical details of Silas\footnote{Download the education version at \url{https://www.depintel.com/silas_download.html}. Note that this version does not include graphical interface and output. Please contact the authors if the reader wishes to generate the figures in Section~\ref{subsec:model-insight} and~\ref{subsec:pred-insight}.}.
%

\section{Machine Learning Preliminaries}
\label{sec:ml}

This section provides the essential definitions of decision trees and their ensembles for classification. The focus is on subtle differences between our implementation and the common definitions in the literature. Specifically, we give a logic-oriented definition of decision trees that facilitates the reasoning and verification of prediction models. We also give the theoretical time complexity of the algorithm training an ensemble of trees in Silas.

\subsection{Decision Trees With a Logical Foundation}
\label{subsec:trees}

In the context of supervised learning, a structured dataset for classification is defined as set of \emph{instances} of the form $\langle x,y \rangle$ where $x = \langle x_1, ..., x_n\rangle$ is an input vector of $n \in \N$ values often called \emph{features} and $y$ is an outcome value often called \emph{label}. We denote by $X$ the feature space and $Y$ the outcome space. In this paper, we focus on binary classification problems where $Y = \{positive,negative\}$.

A decision tree is a tree-like structure composed of internal nodes and terminal nodes called leaves. Internal nodes are predicates over the variables $\{x_1, ..., x_n\}$ corresponding to features. Leaves are sets of instances. Without loss of generality, we focus on binary trees. Internal nodes have two successors respectively called the left and right child nodes. By convention, let $p : X \to \{\top,\bot\}$ be an internal node $v$, the right (resp. left) child node of $v$ is the root of a decision (sub)tree whose set of leaves $L$ is a set of sets of instances such that $\forall~x~\in~\bigcup\{ l~|~l~\in~L\}, p(x) = \top$ (resp. $p(x) = \bot$). It follows that, given a decision tree, any input vector is associated with a single leaf. Further, let $D(Y) : Y \to [0,1]$ be the set of distributions over $Y$ such that $\forall~d~\in~D(Y),\sum_{y~\in~Y}d(y) = 1$. Every given leaf $l$ is associated with a distribution $d_l~\in~D(Y)$ such that for all $y~\in~Y$, $d_l(y)$ is the proportion of instances in $l$ whose outcome is $y$. A decision tree is, therefore, a compact representation of a function of the form $X \to D(Y)$. 

Let $t : X \to D(Y)$ be a tree and $x~\in~X$ be an input vector. Further, let $M : D(Y) \to Y$ be a function such that $\forall~d~\in~D(Y), M(d) = y_{max}$ such that $d(y_{max}) = max\{d(y)~|~y~\in~Y\}$. The outcome predicted by $t$ for the input vector $x$ is the outcome value $M(t(x))$. 

\begin{figure}[ht!]
  \begin{center}
\tikzset{every picture/.style={line width=0.75pt}} 
\begin{tikzpicture}[x=0.75pt,y=0.75pt,yscale=-1,xscale=1]
\draw   (367.47,166) -- (386.19,176.88) -- (367.47,187.77) -- (348.75,176.88) -- cycle ;
\draw   (312.38,226.41) .. controls (312.38,220.4) and (320.76,215.53) .. (331.1,215.53) .. controls (341.44,215.53) and (349.82,220.4) .. (349.82,226.41) .. controls (349.82,232.42) and (341.44,237.3) .. (331.1,237.3) .. controls (320.76,237.3) and (312.38,232.42) .. (312.38,226.41) -- cycle ;
\draw    (367.47,187.77) -- (332.69,214.31) ;
\draw [shift={(331.1,215.53)}, rotate = 322.65] [color={rgb, 255:red, 0; green, 0; blue, 0 }  ][line width=0.75]    (10.93,-3.29) .. controls (6.95,-1.4) and (3.31,-0.3) .. (0,0) .. controls (3.31,0.3) and (6.95,1.4) .. (10.93,3.29)   ;
\draw    (367.47,187.77) -- (401.75,215.36) ;
\draw [shift={(403.3,216.62)}, rotate = 218.82999999999998] [color={rgb, 255:red, 0; green, 0; blue, 0 }  ][line width=0.75]    (10.93,-3.29) .. controls (6.95,-1.4) and (3.31,-0.3) .. (0,0) .. controls (3.31,0.3) and (6.95,1.4) .. (10.93,3.29)   ;
\draw   (403.3,216.62) -- (422.03,227.5) -- (403.3,238.38) -- (384.58,227.5) -- cycle ;
\draw    (403.3,238.38) -- (437.58,265.98) ;
\draw [shift={(439.14,267.23)}, rotate = 218.82999999999998] [color={rgb, 255:red, 0; green, 0; blue, 0 }  ][line width=0.75]    (10.93,-3.29) .. controls (6.95,-1.4) and (3.31,-0.3) .. (0,0) .. controls (3.31,0.3) and (6.95,1.4) .. (10.93,3.29)   ;
\draw    (403.3,238.38) -- (368.52,264.93) ;
\draw [shift={(366.93,266.14)}, rotate = 322.65] [color={rgb, 255:red, 0; green, 0; blue, 0 }  ][line width=0.75]    (10.93,-3.29) .. controls (6.95,-1.4) and (3.31,-0.3) .. (0,0) .. controls (3.31,0.3) and (6.95,1.4) .. (10.93,3.29)   ;
\draw   (348.21,277.03) .. controls (348.21,271.01) and (356.59,266.14) .. (366.93,266.14) .. controls (377.27,266.14) and (385.65,271.01) .. (385.65,277.03) .. controls (385.65,283.04) and (377.27,287.91) .. (366.93,287.91) .. controls (356.59,287.91) and (348.21,283.04) .. (348.21,277.03) -- cycle ;
\draw   (420.42,278.12) .. controls (420.42,272.1) and (428.8,267.23) .. (439.14,267.23) .. controls (449.48,267.23) and (457.86,272.1) .. (457.86,278.12) .. controls (457.86,284.13) and (449.48,289) .. (439.14,289) .. controls (428.8,289) and (420.42,284.13) .. (420.42,278.12) -- cycle ;
\draw (367.47,176.88) node  [align=left] {{\fontfamily{ptm}\selectfont $F_1$}};
\draw (403.3,227.5) node  [align=left] {{\fontfamily{ptm}\selectfont $F_2$}};
\draw (331.1,226.41) node  [align=left] {{\fontfamily{ptm}\selectfont {\footnotesize (0,6)}}};
\draw (366.93,277.03) node  [align=left] {{\fontfamily{ptm}\selectfont {\footnotesize (2,1)}}};
\draw (439.14,278.12) node  [align=left] {{\fontfamily{ptm}\selectfont {\footnotesize (3,1)}}};
\draw (341.99,191.49) node  [align=left] {{\fontfamily{ptm}\selectfont {\footnotesize false}}};
\draw (377.84,242.46) node  [align=left] {{\fontfamily{ptm}\selectfont {\footnotesize false}}};
\draw (396.64,192.24) node  [align=left] {{\fontfamily{ptm}\selectfont {\footnotesize true}}};
\draw (431.44,243.55) node  [align=left] {{\fontfamily{ptm}\selectfont {\footnotesize true}}};
\end{tikzpicture}
  \end{center}
  \caption{An example binary decision tree.}
  \label{fig:tree}
\end{figure}
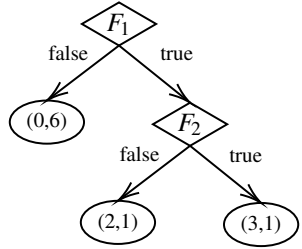

An example of a decision tree is given in Figure~\ref{fig:tree}. In this example, decision nodes are diamonds and leaves are ovals. $F_1$ and $F_2$ are two logical formulae. The pair $(2,1)$ means that there are 3 instances at this leaf, 2 of them are labelled $negative$ and 1 of them is labelled $positive$. 

In Silas, similarly to popular greedy approaches such as C4.5~\cite{Quinlan1993}, trees are constructed by recursively splitting an input dataset until a stopping criterion is satisfied. The splitting predicates are chosen based on the \emph{information gain} they provide, a measure which is computed by comparing the \emph{entropy}~\cite{shannon1948mathematical} between the parent node and the child nodes. Contrary to generic decision trees grown by approaches such as C4.5~\cite{Quinlan1993} the predicates of internal nodes can be arbitrary logical formulae of the propositional logic described below.




A \emph{logical formula} in Silas is defined as an extension of propositional logic with arithmetic terms and comparison operators.
The semantics of the logical language follows that of standard arithmetic and propositional logic. An \emph{arithmetic term} $T$ is defined below where $c$ is a constant (discrete or continuous value) and $var$ is a variable corresponding to (the name of) a  feature:

\begin{center}
$T := c \mid var \mid -T \mid sqrt(T) \mid T + T \mid T - T \mid T * T \mid T / T$
\end{center}

A \emph{Boolean formula} $F$ takes the following form where $C$ denotes a set of constants and $\oplus$ is the exclusive disjunction operator:

\begin{center}
    $F := \top \mid \bot \mid var \in C \mid T < T \mid T \leq T \mid T = T \mid T > T \mid T \geq T \mid$\\ 
    $\lnot F \mid F \land F \mid F \lor F \mid F \limp F \mid F \oplus F$
\end{center}

In the implementation, we use $var\in C$ to express formulae of \emph{nominal} features, which have discrete values and use (in)equalities to express formulae of \emph{numeric} features, which have continuous values.



\subsection{Ensemble of Decision Trees}
\label{subsec:ensemble}

In general, deep decision trees tend to have low bias but high variance due to the fact that they often overfit. Conversely, shallow decision trees tend to have high bias but low variance. To balance bias and variance, a popular approach is to consider multiple trees and aggregate their predictions. In this paper, we focus on \textit{additive ensemble} approaches as they are the most widely adopted.

Let $T = \{\langle w_1, t_1 \rangle, ..., \langle w_m, t_m \rangle\}$ be a set of $m \in \N$ weighted decision trees. We define, similarly to the framework of Cui et al.~\cite{Cui2015}, an additive ensemble $E_T : X \to D(Y)$ as follows. 
\begin{center}
    $\forall~x~\in~X, E_T(x) = \sum_{i=1}^{m} w_i\cdot t_i(x)$
\end{center}

Note that we overload the definition of the addition to apply to distribution over the outcome space as follows.
\begin{center}
    $\forall~d_1,d_2~\in~D(Y), \forall~y~\in~Y, (d_1 + d_2)(y) = \frac{d_1(y)+d_2(y)}{2}$.
\end{center}

In this framework, several popular ensemble methods can be summarised, we briefly describe two of the most popular ones in the sequel.

\paragraph{Bagging} Each decision tree is trained using a subset of the dataset that is sampled uniformly with replacement. The remaining instances form the out-of-bag (OOB) set. The OOB set is often used to measure the performance of trees. When selecting the best formula at each decision node in a tree, it only considers a subset of the features. This is commonly found in algorithms such as Random Forest~\cite{Breiman2001}. Bagging grows large trees with low bias and the ensemble reduces variance.

\paragraph{Boosting} Boosting trains weak learners, i.e., small trees, iteratively as follows:
\begin{center}
    $E_{i+1}(x) = E_i(x) + \alpha_i\cdot t_i(x)$
\end{center}
where $t_i$ is the weak leaner trained at iteration $i$ and $\alpha_i$ is its weight. The final ensemble is thus a special case of $E_T(x)$ above where $w_i$ is $\alpha_i$. Boosting reduces bias.

A well known boosting approach, AdaBoost~\cite{Freund1999}, focuses on training instances that are misclassified in the previous iteration by minimising $\alpha_i$ and $t_i$ in the formula below:
\begin{center}
    $\minimise_{\alpha_i,t_i} \sum_{j=0}^N L(y^{(j)}, E_i(x^{(j)}) + \alpha_i\cdot t_i(x^{(j)}))$
\end{center}
where $L$ is a loss function measuring the difference between the actual outcome $y^{(j)}$ of instance $j$ and $E_{i+1}(x^{(j)})$. AdaBoost often uses exponential loss $L(a,b) = e^{-a\cdot b}$ in which case the shallow decision trees are trained by weighted instances.


\paragraph{Silas} The remainder of the paper is focused on bagging, although the techniques we describe can also be applied to boosting as well as other additive ensemble approaches. In the following, we give the theoretical time complexity of the algorithm training an ensemble of trees in Silas. 


\begin{thm}
Let $t$ be the number of trees, $m$ be the number of attributes and $N$ be the number of data points. The complexity of training an ensemble of trees in Silas is $O(t*m*N^2*log(N))$ in the worst case, and $O(t*\sqrt{m}*N*log(N))$ on average.
\end{thm}

\begin{pf} (Outline)
Silas builds trees by recursively splitting leaves. Given $n$ data points, the complexity of finding an appropriate split predicate is $O(n)$ for a numerical attribute and $O(n + c*log(c))$ for a nominal attribute with $c$ distinct values. This operation has to be performed at each of the internal nodes of the tree. Further, we note that multiple attributes are considered at each splitting nodes.

In the worse case, all attributes are nominal with $N$ distinct values, and there are $t$ trees with $N$ leaves, i.e., one data point per leaf. This corresponds to the computation of $t*(2*N+1)$ splitting predicates. Assuming that all attributes are considered at each split, it follows that the complexity of training an ensemble of tree in Silas is $O(t*m*N^2*log(N))$ in the worst case.

In the average case, nominal attributes have at most $c$ distinct values such that $c \ll N$. It follows that the complexity of finding an appropriate split predicate is $O(N)$ for both nominal and numerical attributes. Further, only $\sqrt{m}$ attributes are considered at each split, and trees have $log(N)$ leaves on average. Hence, growing $t$ trees corresponds to the computation of $t*(2*log(N)+1)$ splitting predicates. It follows that the complexity of training an ensemble of tree in Silas is, in the average case, $O(t*\sqrt{m}*N*log(N))$.

    We note that these complexity results are similar to those of the random forest algorithm~\cite{louppe2014}.
\qed
\end{pf}



\section{Proposed Explainable And Verifiable Machine Learning}
\label{sec:reason}

This section is concerned with eXplainable AI and safe machine learning. We describe our solution towards a more trustworthy machine learning technique using logic and automated reasoning as the backbone. We discuss the Model Audit module for formally verifying prediction models against user specifications, the Enforcement Learning module for training correct-by-construction models, Model Insight for explaining prediction models and Prediction Insight for explaining prediction instances.

\subsection{Logical Semantics of Decision Trees}
\label{subsec:extract}

Given a decision tree, we can obtain the following types of logical formulae: 
\begin{description}
\item[Internal Node formula:] The logical formula corresponding to every internal node.
\item[Branch formula:] The logical formula $(\bigwedge N) \limp (y = M(d_l))$, where $N$ is the set of internal node formulae along the branch leading to the leaf $l$ and $d_l$ is the distribution associated with $l$. 
\item[Tree formula:] The logical formula $\bigvee B$ where $B$ is the set of branch formulae. 
\end{description}

Each of the above mentioned formula is associated with a weight. An internal node formula is weighted by the information gain computed during training. A branch formula leading to a leaf $l$ is weighted by the value $log_2(2) - H(l)$ where $H(l)$ is the entropy~\cite{shannon1948mathematical} of the leaf $l$. A tree formula is weighted by the ROC-AUC score obtained on its out-of-bag sample during training.



\subsection{Model Audit}
\label{subsec:audit}

The purpose of the model audit module is to provide the means to formally certify that the prediction model complies with user specifications. To do so we adopt advanced automated reasoning techniques, especially satisfiability modulo theories (SMT) solvers~\cite{de2011satisfiability}. SMT solvers determine the \textit{satisfiability} of logical formulae with respect to combinations of background theories expressed in classical first-order logic with equality. A logical formula $f$ is said satisfiable, denoted by $SAT(f)$, if and only if there exists a valuation assigning values to its variables such that it is evaluated to true. 

A \emph{user specification} is a tuple $S = \langle s_\bot, s_\top \rangle$ where $s_\bot$ and $s_\top$ are logical formula over $\{x_1, ..., x_n\}$. A prediction model complies with the user specification $S$ if for all input instance $x~\in~X$ leading to a $positive$ (resp. $negative$) prediction, $s_\top(x)$ (resp. $s_\bot(x)$) evaluates to true. Formally, a user specification $S$ is valid over a prediction model $G : X \to D(Y)$, denoted by $G \models S$, if and only if:
\begin{center}
    $\forall~x~\in~X, ((M(G(x)) = negative) \limp s_\bot(x)) \wedge ((M(G(x)) = positive) \limp s_\top(x))$.
\end{center}

In order to verify the validity of a user specification $S$ over an ensemble trees model $E_T$, i.e. $E_T \models S$, we propose to reduce the problem to the verification of the validity of $S$ over each of the decision trees in $T$. The following theorem proves the soundness of this approach.

\begin{thm}[Soundness]\label{soundness}
If $S$ is valid over all tree in $T$, i.e. $\forall~t_i~\in~T, t_i \models S$, then $S$ is valid over $E_T$, i.e. $E_T \models S$.
\end{thm}
\begin{pf} (Outline)
    Without loss of generality, let us consider an ensemble trees model $E_T$ based on two trees, i.e. $T = \{\langle w_1, t_1 \rangle , \langle w_2, t_2 \rangle\}$. Now assume that $\forall~t_i~\in~T, t_i \models S$. 
    Given an arbitrary $x \in X$, we have to consider two case: (i) $M(t_1(x)) = M(t_2(x))$ and (ii) $M(t_1(x)) \neq M(t_2(x))$.
    \textbf{Case i:} Let $y \in Y$ such that $M(t_1(x)) = M(t_2(x)) = y$ then, by definition of $E_T$, we have $M(E_T(x)) = y$ and we can conclude that $((M(E_T(x)) = negative) \limp s_\bot(x)) \wedge ((M(E_T(x)) = positive) \limp s_\top(x))$ holds.
    \textbf{Case ii:} Without loss of generality, let us consider the case where $M(t_1(x)) = positive$ and $M(t_2(x)) = negative$. Since we assumed that $t_1 \models S$ and $M(t_1(x)) = positive$, we know that $s_\top(x)$ holds. Similarly, since we assumed that $t_2 \models S$ and $M(t_2(x)) = negative$, we know that $s_\bot(x)$ holds. We can conclude that $s_\top(x) \wedge s_\bot(x)$ holds, hence $((M(E_T(x)) = negative) \limp s_\bot(x)) \wedge ((M(E_T(x)) = positive) \limp s_\top(x))$ holds.
    \qed
\end{pf}

We note that this reduction is not sound when considering multiclass classification where the number of classes is greater than two. Further, this reduction is not complete since a tree could violate the specification while being outnumbered by trees that comply with the specification in the aggregation phase of the ensemble tree model. It follows that verification procedures based this reduction are therefore incomplete. This is a trade-off purposefully made to reduce the overall complexity in order to achieve better scalability.

We now proceed to show that, using the reduction we described, SMT solver can be efficiently applied to the verification of user specification over ensemble tree models.
Let $t$ be a tree and $F_t$ its corresponding tree formula, the user specification $S$ is valid over $t$ if and only if:
\begin{center}
    $\neg SAT(F_t \wedge \neg (((y = negative) \limp s_\bot(x)) \wedge ((y = positive) \limp s_\top(x))))$
\end{center}

By Theorem~\ref{soundness} this means that we can use SMT solvers to verify the validity of a user specification over ensemble trees models. This can be done in a parallel fashion since each tree of an ensemble tree model can be verified independently. 

\paragraph{Silas} The Model Audit feature employs the Z3 solver~\cite{deMoura2008}. The interaction with Z3 is straightforward as Silas supports direct translation from logical formulae to the \texttt{z3::expr} type in Z3 C++ binding. 

The remainder of this section presents a case study and report experimental results demonstrating the feasibility and efficiency of the proposed approach.

\paragraph{Case study} We use the Kick dataset as a real-life application to illustrate the Model Audit feature\footnote{Detailed case study can be found at \url{https://www.depintel.com/documentation/_build/html/tutorials/advanced.html}}. The goal of this dataset is to predict whether a used car at an auction is a good buy or a bad buy. In a hypothetical scenario where car maker $B$ discovered that model $C$ produced in year $YY$ have problems and they had recalled all those cars. We wish to check if our prediction model already ``knows'' this. We formulate the specification as follows: $(y = positive) \limp \lnot(make = B \land model = C \land year = YY)$. The Model Audit feature can be used to check if the prediction model meets the specification. In case it does not, we can use Enforcement Learning described in the following section to train a new model that builds in this information. Running Model Audit on the new model again shows that it meets the specification (guaranteed). 

\begin{figure}[t!]
\begin{center}
\centering
\includegraphics[width=0.65\textwidth]{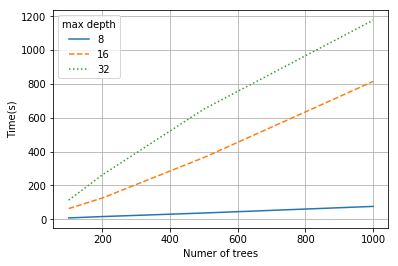}
\caption{Experiment results of Model Audit.}
\label{fig:exp-model-audit}
\end{center}
\end{figure}

To evaluate the efficiency of the verification procedure, we generate models of various sizes (in terms of the number of trees and leaf size) for the Kick dataset and record the computation time of the Model Audit feature when verifying the above property. Experimental results are given in Figure~\ref{fig:exp-model-audit}. We observe that, as expected, the computation time grows linearly with respect to the number of trees and exponentially with respect to the depth of trees. Full trees in this example are often smaller than depth 32, so the increase from depth 16 to 32 is not large. The verification time for models with positive results and negative results are almost identical. Overall, the verification can be done in a reasonable time ($<$ 20 min) for models with 1000 trees of depth 32.

\subsection{Enforcement Learning}
\label{sec:enforce}

The purpose of the \textit{Enforcement Learning} module is to provide the means to build prediction models that, given a user specification, are correct-by-construction. This feature is notably used in the context of critical or regulated applications. It can also be used to enforce additional knowledge given by domain experts or existing expert systems. 

Given a user specification $S = \langle s_\bot, s_\top \rangle$, Enforcement Learning proceeds as follows:

(1) It filters out from the dataset all instance $\langle x,y \rangle$ where:
\begin{center}
    $((y=negative) \wedge \neg(s_\bot)) \vee ((y=positive) \wedge \neg(s_\top))$
\end{center}

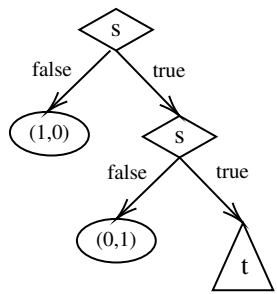
\begin{figure}[h!]
  \begin{center}

\tikzset{every picture/.style={line width=0.75pt}} 

\begin{tikzpicture}[x=0.75pt,y=0.75pt,yscale=-1,xscale=1]

\draw   (286.5,9) -- (305,19.5) -- (286.5,30) -- (268,19.5) -- cycle ;
\draw   (232.99,72) .. controls (232.37,65.92) and (240.78,61) .. (251.75,61) .. controls (262.73,61) and (272.12,65.92) .. (272.74,72) .. controls (273.35,78.08) and (264.95,83) .. (253.97,83) .. controls (242.99,83) and (233.6,78.08) .. (232.99,72) -- cycle ;
\draw   (350.5,117) -- (365.75,151) -- (335.25,151) -- cycle ;
\draw    (286.5,30) -- (317.11,61.56) ;
\draw [shift={(318.5,63)}, rotate = 225.88] [color={rgb, 255:red, 0; green, 0; blue, 0 }  ][line width=0.75]    (10.93,-3.29) .. controls (6.95,-1.4) and (3.31,-0.3) .. (0,0) .. controls (3.31,0.3) and (6.95,1.4) .. (10.93,3.29)   ;

\draw    (283.63,30) -- (253.19,59.61) ;
\draw [shift={(251.75,61)}, rotate = 315.8] [color={rgb, 255:red, 0; green, 0; blue, 0 }  ][line width=0.75]    (10.93,-3.29) .. controls (6.95,-1.4) and (3.31,-0.3) .. (0,0) .. controls (3.31,0.3) and (6.95,1.4) .. (10.93,3.29)   ;

\draw   (318.5,63) -- (337,73.5) -- (318.5,84) -- (300,73.5) -- cycle ;
\draw   (266.75,126) .. controls (266.75,119.92) and (275.65,115) .. (286.63,115) .. controls (297.6,115) and (306.5,119.92) .. (306.5,126) .. controls (306.5,132.08) and (297.6,137) .. (286.63,137) .. controls (275.65,137) and (266.75,132.08) .. (266.75,126) -- cycle ;
\draw    (318.5,84) -- (288.06,113.61) ;
\draw [shift={(286.63,115)}, rotate = 315.8] [color={rgb, 255:red, 0; green, 0; blue, 0 }  ][line width=0.75]    (10.93,-3.29) .. controls (6.95,-1.4) and (3.31,-0.3) .. (0,0) .. controls (3.31,0.3) and (6.95,1.4) .. (10.93,3.29)   ;

\draw    (318.5,84) -- (349.11,115.56) ;
\draw [shift={(350.5,117)}, rotate = 225.88] [color={rgb, 255:red, 0; green, 0; blue, 0 }  ][line width=0.75]    (10.93,-3.29) .. controls (6.95,-1.4) and (3.31,-0.3) .. (0,0) .. controls (3.31,0.3) and (6.95,1.4) .. (10.93,3.29)   ;

\draw (286.5,19.5) node [scale=1] [align=left] {s};
\draw (252.86,72) node [scale=0.8] [align=left] {(1,0)};
\draw (286.63,126) node [scale=0.8] [align=left] {(0,1)};
\draw (351.5,139) node [scale=1] [align=left] {t};
\draw (313.25,40.5) node [scale=0.8] [align=left] {true};
\draw (345.25,91.5) node [scale=0.8] [align=left] {true};
\draw (254.25,39.5) node [scale=0.8] [align=left] {false};
\draw (292.25,91.5) node [scale=0.8] [align=left] {false};
\draw (318.5,73.5) node [scale=1] [align=left] {s};

\end{tikzpicture}

  \end{center}
  \caption{Template of correct by construction trees}
  \label{enforcedTree}
\end{figure}

(2) It constructs trees of the form given by figure~\ref{enforcedTree} where $t$ is a tree grown from the filtered dataset.

Trees built according to the above procedure are, by construction, valid with respect to the given user specification. By theorem~\ref{soundness} the resulting ensemble tree models are also valid with respect to the given user specification.

\subsection{Model Insight}
\label{subsec:model-insight}

When the user obtains a model with satisfactory performance, we provide a feature named Model Insight for analysing the general decision-making of the model. 

Given a label $v$ (e.g. $positive$), we are interested in knowing which set of input values would be predicted as $v$ by an ensemble tree model $E_T$. This way, domain experts may use their knowledge to confirm or refute the rationale exhibited by $E_T$ when predicting label $v$.

To achieve this, we consider the set $B$ of branches in trees of $E_T$ that predict $v$. This set corresponds to the set $F_B$ of weighted branch formulae. We can then apply automated reasoning techniques to extract the maximum satisfiable subset corresponding to the set of input values on which the majority of branches in $B$ agree. This subset is called the max-sat core (MSC). We can do so using SMT solvers such as Z3~\cite{deMoura2008}. However, when considering all formulae in $F_B$, the resulting MSC relates to a very small set of input values for which the model predicts $v$ with very high confidence. Such an MSC corresponds to very specific cases that are not useful in general explanations of the prediction model. Therefore, to broaden the scope of the explanation we sample at three different levels: the node level, the branch level and the tree level. The sampling results in MSCs that correspond to more general explanations. 

The explanations based on MSC extraction can be combined with feature importance to better illustrate the decision-making of the prediction model. There are several methods to compute feature importance in the literature, e.g.,~\cite{Altmann2010}. Our computation and presentation of feature importance are inspired by the LIME tool~\cite{lime2016} and the SHAP method~\cite{Lundberg2017}.

The remainder of this section presents a small case study that illustrates the above approach.

\begin{figure}[h!]
\begin{floatrow}
\ffigbox{%
  \includegraphics[width=0.6\textwidth]{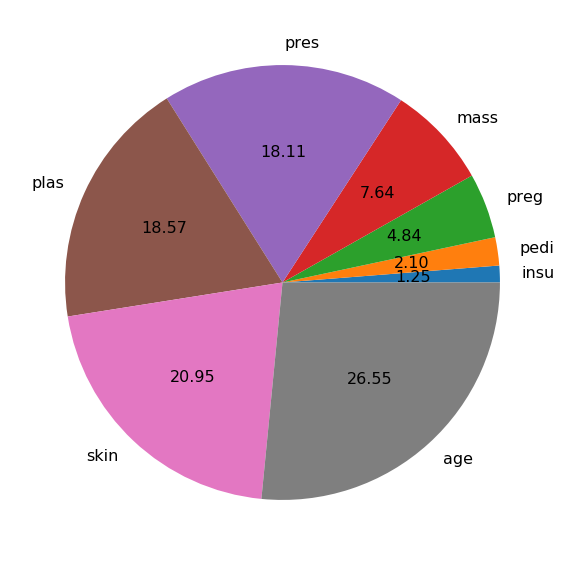}
}{%
  \caption{Model Insight: feature importance.}%
  \label{fig:model-insight}
}\hspace{30px}
\capbtabbox{%
\begin{small}
  \begin{tabular}{ c C{3cm} }
        \hline
        & Decision Logic \\[2px]
        \hline
        \parbox[t]{2mm}{\multirow{5}{*}{\rotatebox[origin=c]{90}{Positive}}} & $30 \leq $ {\color{gray} age} $< 47$\\
        & $31\leq $ {\color{magenta} skin} $< 99$\\
        & $155 \leq$ {\color{olive} plas} $< 157$\\
        & $40 \leq$ {\color{violet} pres} $< 122$\\
        & $30 \leq$ {\color{red} mass} $< 40.8$\\[5px]
        \hline
  
        \parbox[t]{2mm}{\multirow{5}{*}{\rotatebox[origin=c]{90}{Negative}}} & $21 \leq $ {\color{gray} age} $< 27$\\
        & $0 \leq$ {\color{magenta} skin} $< 31$\\
        & $103 \leq$ {\color{olive} plas} $< 120$\\
        & $0 \leq$ {\color{violet} pres} $< 68$\\
        & $0 \leq$ {\color{red} mass} $< 29.8$\\
        \hline
\end{tabular}
\end{small}
\vspace{20px}
}{%
  \caption{Model Insight: decision logic.}%
  \label{tab:model-insight}
}
\end{floatrow}
\end{figure}

\paragraph{Case study} Consider the diabetes dataset~\cite{diabetes}. The eight features are the number of times pregnant (preg), plasma glucose concentration (plas), diastolic blood pressure (pres), 2-hour serum insulin (insu), triceps skinfold thickness (skin), body mass index (mass), diabetes pedigree function (pedi) and age. We build a forest of 100 trees with the default settings of Silas and perform the Model Insight analysis on the best model in 10-fold cross-validation. Figure~\ref{fig:model-insight} shows the feature importance score of the model. The values are normalised into percentages, thus, we can read the figure as ``the feature age contributes 26.55\% of the decision making of the model''. Table~\ref{tab:model-insight} gives the general decision logic of the same model. The pedi and insu features are less important and we do not show them in Table~\ref{tab:model-insight}. The decision logic is divided into the constraints that lead to positive diabetes and those that lead to negative diabetes. Medical practitioners can cross-reference the pie chart and the table to evaluate whether the logic of the model is consistent with their knowledge. Disclaimer: the diabetes dataset only contains 768 instances and their characteristics may not be representative for a larger population.

\subsection{Prediction Insight}
\label{subsec:pred-insight}

The prediction insight aims at providing users with the decision logic corresponding to individual predictions. This aspect is often an essential element in critical or regulated predictive applications.  

Any decision tree, and by extension any ensemble tree model, can be seen as a simple decision rules system composed of rules of the form:
\begin{centering}
     \texttt{if} condition \texttt{then} prediction.
\end{centering}

Consider an instance $x \in X$ and the decision tree $t$, the condition of the decision rule associated with the prediction $t(x)$ is the branch formula that leads to the prediction $t(x)$. Likewise, consider the ensemble tree $E_T$, the condition of the decision rule associated with the prediction $E_t(x)$ is the conjunction of all the branch formula that lead to the predictions $t_1(x), ..., t_m(x)$.

Similar to model insights, prediction insights can be mixed with feature importance scores of individual predictions obtained from feature attribution methods such as SHAP~\cite{Lundberg2017} as illustrated by the following case study. 




\begin{figure}[h!]
  \centering
  \includegraphics[width=\textwidth]{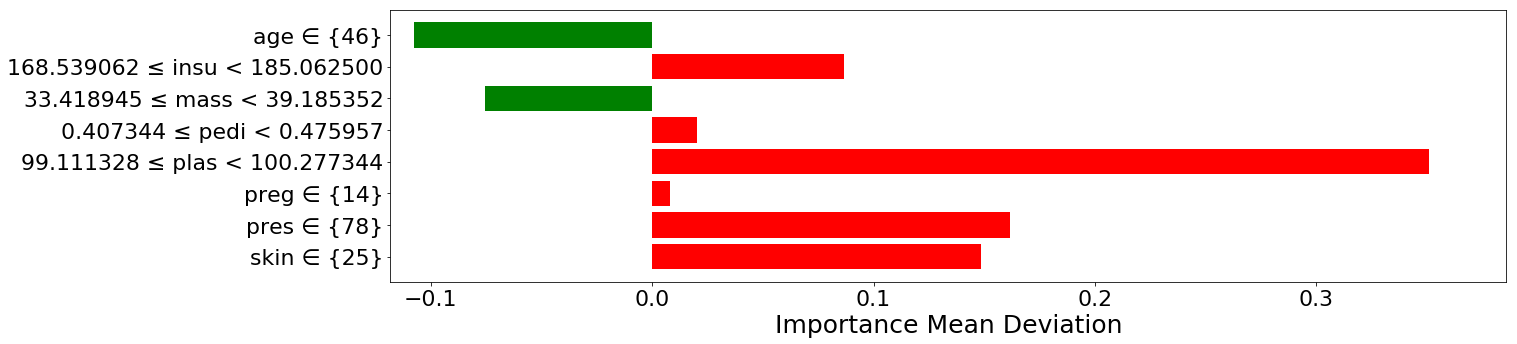}
  \caption{Prediction Insight examples.}\label{fig:pred-insight}
\end{figure}

\paragraph{Case study} Figure~\ref{fig:pred-insight} illustrates a typical prediction insight's output on an instance from the diabetes dataset. The model predicts that there is 62.96\% chance that the patient has diabetes. The figure shows how each feature contributes to the prediction and the range at which it does so.

\section{Performance Results and Analysis}
\label{sec:comp}

This section provides experimental results on the predictive, time and memory performances of Silas. It also describes relevant implementation details and lesson learned during the development. 

There are multiple incentives to develop high-performance machine learning tools. The benefits include a smaller ecological footprint as well as cost savings and increase in productivity. Reducing the hardware requirements of machine learning applications also fosters security as the data no longer need to be transmitted through off-site cloud infrastructures and can instead be locally hosted.  



\subsection{Comparative Results On Predictive Performance}
\label{subsec:perf-exp}

We show empiric evidence that our implementation of ensemble trees is fast, efficient and has state-of-the-art \emph{predictive performance}.

\begin{table}[ht!]
\footnotesize
\begin{tabular}{ c c c c c c c }
        \hline
        Dataset & \#Instances & \#Numeric & \#Nominal & Max Norminal & \#Missing  & Balance\\
         & & Features & Features & Cardinality & Values & \\
        \hline
        Kick~\cite{kick} & 72,983 & 14 & 18 & 1,063 & 149,271 & 7:1\\
        Creditcard~\cite{creditcard} & 284,807 & 30 & 0 & N.A. & 0 & 577:1\\
        Flight~\cite{flight} & 10M & 2 & 6 & 315 & 0 & 4:1\\
        Higgs~\cite{higgs} & 10.5M & 28 & 0 & N.A. & 0 & 1:1\\
        \hline
\end{tabular}
\caption{Details of datasets.}
\label{tab:datasets}
\end{table}

\begin{figure}[ht!]
\begin{subfigure}{.5\textwidth}
  \centering
  \includegraphics[width=\linewidth,height=5.6cm]{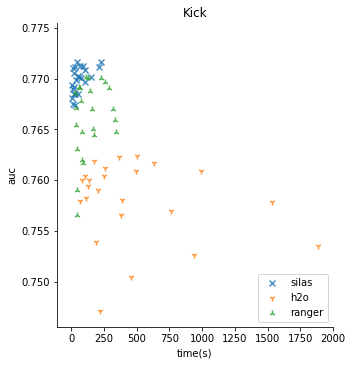}
  \label{fig:sfig1}
\end{subfigure}%
\begin{subfigure}{.5\textwidth}
  \centering
  \includegraphics[width=\linewidth,height=5.6cm]{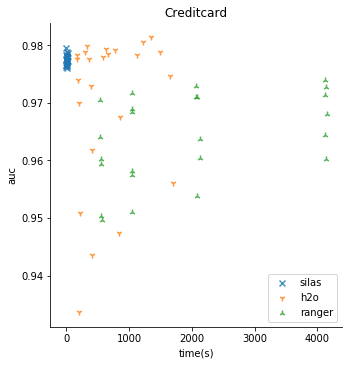}
  \label{fig:sfig2}
\end{subfigure}
\vspace{-0.5cm}
\begin{subfigure}{.5\textwidth}
  \centering
  \includegraphics[width=\linewidth,height=5.6cm]{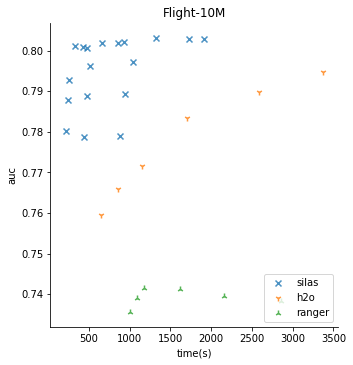}
  \label{fig:sfig1}
\end{subfigure}%
\begin{subfigure}{.5\textwidth}
  \centering
  \includegraphics[width=\linewidth,height=5.6cm]{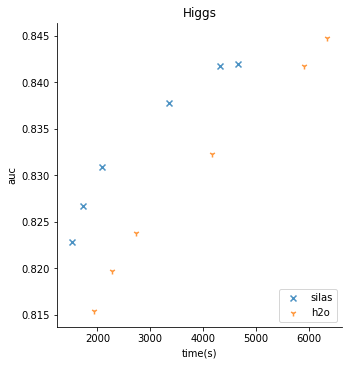}
  \label{fig:sfig2}
\end{subfigure}
\vspace{-0.3cm}
\caption{Experimental results plotted by time(s) and area under the ROC curve (AUC).}
\label{fig:exp}
\end{figure}

We compared Silas with H2O~\cite{h2o} and Ranger~\cite{Wright2015}, which are industry leaders of random forest implementations. The results reported in this paper were obtained on a desktop machine with an Intel Core i7-7700 processor and 32 GB of memory. We selected four datasets~\cite{kick,creditcard,flight,higgs}, shown in Table~\ref{tab:datasets}, with the following criteria: (1) They have at least 50,000 instances. (2) They are public datasets. (3) They are binary classification problems. (4) They are datasets with real-life applications. We focus on these datasets in order to study how different parameters affect the predictive performance for each dataset. The chosen datasets have a variety of characteristics in terms of missing values, types of features, the balance of classes, etc., and should reflect the tools' abilities in different scenarios. Notably, we chose Kick and Flight because they have a mixture of numeric features and nominal features, and the nominal feature with the largest number of unique values, indicated by ``Max Norminal Cardinality'' in Table~\ref{tab:datasets}, has hundreds of values. 


For each dataset, we ran the tools with \emph{all combinations} of the following settings:
\textit{Number of trees}: 100, 200, 400, 800. \textit{Number of instances at leaves (leaf size)}: 1, 4, 16, 64, 128, 256. \textit{Max depth of trees}: 64 or unlimited.
The other parameters are set to default~\cite{silasparameters,h2oparameters,rangerparameters}. The training time is limited to 2 hours. The results are shown in Figure~\ref{fig:exp}. Each point in the figure corresponds to a model with a certain number of trees and leaf size. The vertical axis reports the ROC-AUC obtained using 10-fold cross-validation (Kick and Creditcard) or on a test dataset (Flight and Higgs). The horizontal axis reports the time (in seconds) of the computation including the training phase as well as the time required to compute the ROC-AUC metric. For instance, the six points for Silas in the Higgs sub-figure are models of 100 trees and leaf size 256 to 1 respectively (left to right). In the Flight-10M sub-figure, three ``columns'' of points are observed for Silas. Those correspond to the models of 100 trees, 200 trees and 400 trees respectively (left to right).

\paragraph{Observations} Generally speaking, the points for Silas are more clustered than other tools, which indicates that Silas's results are less sensitive to hyperparameter settings. For Kick, Silas and Ranger outperformed H2O in a much shorter time. For Creditcard, Silas and H2O slightly outperformed Ranger, and Silas is much faster than the other two. For Flight, Silas outperformed both H2O and Ranger and is the only tool that can generate 200 and 400 trees within the time limit and without crashing. For Higgs, Silas and H2O yielded similar results with 100 trees, however, Silas is faster. Ranger failed to parse the dataset.

\paragraph{Synthesis} Overall, we observe that Silas trains models that are as good as H2O and Ranger in a much shorter time. The predictive performance of Silas is very consistent across different parameters and datasets and often better than comparable tools. We note, however, that H2O slightly outperformed Silas under some parameters on Creditcard and Higgs. This can be explained by the fact that the search resolution over numerical features is fixed in Silas whereas the resolution is adaptive in H2O. In future work, we plan on improving the discretisation of numerical features in order to improve predictive performance.

\subsection{Implementation}

We list some of the major implementation and design choices which we believe contribute to the excellent time and memory efficiency of Silas.

Efficient software is built on a good foundation. We have studied various programming languages and styles and settled with the following three key points at the core of Silas code base. 

\paragraph{Programming language} The programming language itself must be efficient and low-level enough to give us the liberty to perform cache and instruction level optimisations. According to recent benchmarks for comparing the speed of programming languages, e.g.,~\cite{heergithub}, Go, C, C++ and Crystal are among the fastest languages. We chose C++ because it also provides enough high-level programming features, such as template metaprogramming~\cite{Abrahams2004}, which are useful when realising the other key points.

\paragraph{Programming paradigm} We adopt a programming paradigm largely inspired by the functional programming paradigm. More specifically, we predominantly employ \emph{pure functions} due to the following beneficial properties~\cite{Carmack2012}: reusability, testability, thread safety and the absence of side effects. To do so, we developed a C++ framework that enables us to statically compose pure functions in a straight forward manner. This framework notably relies on a static dispatch technique called Curiously Recurring Template Pattern (CRTP)~\cite{Abrahams2004} to offer efficient means of sequential and parallel compositions. In a multi-core execution environment, this organisational framework incurs no run-time overhead.

\paragraph{Programming design} Our codebase follows the data-oriented design approach. The emphasis is placed on the data being created, manipulated and stored. The main advantage of data-oriented programs is the constraint on the locality of reference which enables safe and effective parallelism (e.g. vectorisation of code). Another benefit of data-oriented programming is the efficient use of memory caching, an essential aspect of modern hardware.\\

Like many high-performance tools, the development of Silas required multiple attempts. There were two major recodings. We first coded Silas as a classic C++ object-oriented application. This version, referred to as the \textit{legacy version}, served as the baseline of the subsequent implementation. Due to the difficulty we had in safely parallelising the core algorithms of Silas (e.g. training, computation of predictive metrics) we decided to adopt a pure-function based framework inspired by functional programming aspects. This second implementation, referred to as the \textit{$1^{st}$ revision}, was more stable and efficient than the legacy version. It also forced us to reconsider the data structures in use. For instance, we switched to using a column-based database instead of the more traditional row-based database. Finally, in order to further optimise our code base (e.g. optimise caching, increase information density), we made the design choice to adopt strongly data-oriented coding guidelines. This lead to a new implementation of Silas, referred to as the \textit{$2^{nd}$ revision}. 

\begin{table}[h!]
    \centering
\begin{tabular}{llllll}
\hline
\multicolumn{6}{l}{1M flights Dataset}                                                                        \\ \hline
Metrics           & Legacy & $1^{st}$ Rev                         & $2^{nd}$ Rev                           & H2O   & Ranger \\ \hline
Time (s)          & 77.5   & 23.1                          & \textbf{16.8}  & 61.2  & 62.9   \\
Memory Usage (GB) & 3      & \textbf{0.6} & \textbf{0.6}   & 5.1   & 3.1    \\
ROC-AUC               & 0.757  & 0.753                         & 0.752                           & 0.744 & 0.722  \\ \hline
\multicolumn{6}{l}{10M flights Dataset}                                                                       \\ \hline
Metrics           & Legacy & $1^{st}$ Rev                         & $2^{nd}$ Rev                           & H2O   & Ranger \\
\hline
Time (s)          & 804.4  & 281.1                         & \textbf{195.7} & 1216  & 1195.9 \\
Memory Usage (GB) & 25.9   & 4.6                           & \textbf{4.3}   & 12.6  & 30     \\
ROC-AUC               & 0.802  & 0.796                         & 0.793                           & 0.772 & 0.741  \\ \hline
\end{tabular}
    \caption{Computational performance for the Flight dataset. We report the \emph{total time} used for loading data, training, testing and computing the predictive performance. The Memory Usage row shows the \emph{peak memory usage} during the computation.}
    \label{tab:time-space-perf}
\end{table}

To illustrate the evolution of Silas throughout its recoding phases, Table~\ref{tab:time-space-perf} shows results obtained over the Flight dataset~\cite{szilardbenchmark} using the legacy, $1^{st}$ revision and $2^{nd}$ revision of Silas as well as ranger and H2O for reference. We can observe that the time efficiency of Silas significantly improved with each revision. We also note that $1^{st}$ revision of Silas significantly improved its memory efficiency. More specifically, the $2^{nd}$ revision of Silas is 3.6x faster than H2O on the 1M dataset and is 6.1x faster than Ranger on the 10M dataset. In terms of memory usage, Ranger uses 5x more memory than the $2^{nd}$ revision of Silas on the 1M dataset, and H2O uses 3x more memory than Rev 2 on the 10M dataset.

\section{Conclusion And Future Work}
\label{sec:conc}

This work proposed a new classification tool called Silas, which has state-of-the-art predictive performance and often runs faster and uses less memory than other implementations. Moreover, Silas provides features for a more dependable machine learning service. These include: Model Audit, which formally verifies user specifications against predictive models; Enforcement Learning, which generates predictive models that are guaranteed to satisfy user specifications; Model Insight, which provides explanations on how the model works; and a special case of the above called Prediction Insight, which explains how a particular prediction is made. 

Although not discussed in this paper, Silas does support multi-class classification. However, we have not investigated how verification and explanation analysis can be done for multiple classes. Similarly, we have not studied how to make regression tasks more white-box. These topics will be our focus of future research and development directions.

\bibliographystyle{elsarticle-num}
\bibliography{main}

\begin{thebibliography}{10}
\expandafter\ifx\csname url\endcsname\relax
  \def\url#1{\texttt{#1}}\fi
\expandafter\ifx\csname urlprefix\endcsname\relax\def\urlprefix{URL }\fi
\expandafter\ifx\csname href\endcsname\relax
  \def\href#1#2{#2} \def\path#1{#1}\fi

\bibitem{Gomez-Uribe2015}
C.~A. Gomez-Uribe, N.~Hunt, \href{http://doi.acm.org/10.1145/2843948}{The
  netflix recommender system: Algorithms, business value, and innovation}, ACM
  Trans. Manage. Inf. Syst. 6~(4) (2015) 13:1--13:19 (Dec. 2015).
\newblock \href {https://doi.org/10.1145/2843948} {\path{doi:10.1145/2843948}}.
\newline\urlprefix\url{http://doi.acm.org/10.1145/2843948}

\bibitem{Eliot2017}
L.~Eliot, M.~Eliot, Autonomous Vehicle Driverless Self-Driving Cars and
  Artificial Intelligence: Practical Advances in AI and Machine Learning, 1st
  Edition, LBE Press Publishing, 2017 (2017).

\bibitem{Turkson2016}
R.~E. {Turkson}, E.~Y. {Baagyere}, G.~E. {Wenya}, A machine learning approach
  for predicting bank credit worthiness, in: 2016 Third International
  Conference on Artificial Intelligence and Pattern Recognition (AIPR), 2016,
  pp. 1--7 (Sep. 2016).
\newblock \href {https://doi.org/10.1109/ICAIPR.2016.7585216}
  {\path{doi:10.1109/ICAIPR.2016.7585216}}.

\bibitem{Kononenko2001}
I.~Kononenko, \href{http://dx.doi.org/10.1016/S0933-3657(01)00077-X}{Machine
  learning for medical diagnosis: History, state of the art and perspective},
  Artif. Intell. Med. 23~(1) (2001) 89--109 (Aug. 2001).
\newblock \href {https://doi.org/10.1016/S0933-3657(01)00077-X}
  {\path{doi:10.1016/S0933-3657(01)00077-X}}.
\newline\urlprefix\url{http://dx.doi.org/10.1016/S0933-3657(01)00077-X}

\bibitem{cumby2004}
C.~Cumby, A.~Fano, R.~Ghani, M.~Krema, Predicting customer shopping lists from
  point-of-sale purchase data, in: Proceedings of the tenth ACM SIGKDD
  international conference on Knowledge discovery and data mining, ACM, 2004,
  pp. 402--409 (2004).

\bibitem{nypost2016}
N.~Y. Post, Toddler asks amazon’s alexa to play song but gets porn instead,
  \url{https://nypost.com/2016/12/30/toddler-asks-amazons-alexa-to-play-song-but-gets-porn-instead/}
  (2016).

\bibitem{guardian2018}
T.~Guardian, Self-driving uber kills arizona woman in first fatal crash
  involving pedestrian,
  \url{https://www.theguardian.com/technology/2018/mar/19/uber-self-driving-car-kills-woman-arizona-tempe}
  (2018).

\bibitem{bonacina2017automated}
M.~P. Bonacina, Automated reasoning for explainable artificial intelligence,
  in: ARCADE Workshop (in association with CADE-26), Gothenburg, Sweden, 2017
  (2017).

\bibitem{szilardbenchmark}
S.~Pafka, A minimal benchmark for scalability, speed and accuracy of commonly
  used open source implementations of the top machine learning algorithms for
  binary classification, https://github.com/szilard/benchm-ml (2018).

\bibitem{Chen2016}
T.~Chen, C.~Guestrin,
  \href{http://doi.acm.org/10.1145/2939672.2939785}{Xgboost: A scalable tree
  boosting system}, in: Proceedings of the 22Nd ACM SIGKDD International
  Conference on Knowledge Discovery and Data Mining, KDD '16, ACM, New York,
  NY, USA, 2016, pp. 785--794 (2016).
\newblock \href {https://doi.org/10.1145/2939672.2939785}
  {\path{doi:10.1145/2939672.2939785}}.
\newline\urlprefix\url{http://doi.acm.org/10.1145/2939672.2939785}

\bibitem{h2o}
H2O, H2o, https://github.com/h2oai/h2o-3 (2019).

\bibitem{Wright2015}
M.~Wright, A.~Ziegler, ranger: A fast implementation of random forests for high
  dimensional data in c++ and r, Journal of Statistical Software 77 (08 2015).
\newblock \href {https://doi.org/10.18637/jss.v077.i01}
  {\path{doi:10.18637/jss.v077.i01}}.

\bibitem{Bride18}
H.~Bride, J.~Dong, J.~S. Dong, Z.~H{\'{o}}u, Towards dependable and explainable
  machine learning using automated reasoning, in: Formal Methods and Software
  Engineering - 20th International Conference on Formal Engineering Methods,
  {ICFEM} 2018, Gold Coast, QLD, Australia, November 12-16, 2018, Proceedings,
  2018, pp. 412--416 (2018).

\bibitem{depintel}
D.~Intelligence, Silas, \url{https://www.depintel.com/silas_about.html} (2019).

\bibitem{lime2016}
M.~T. Ribeiro, S.~Singh, C.~Guestrin, "why should {I} trust you?": Explaining
  the predictions of any classifier, in: Proceedings of the 22nd {ACM} {SIGKDD}
  International Conference on Knowledge Discovery and Data Mining, San
  Francisco, CA, USA, August 13-17, 2016, 2016, pp. 1135--1144 (2016).

\bibitem{Hara2016}
S.~Hara, K.~Hayashi, Making tree ensembles interpretable (06 2016).

\bibitem{Lundberg2017}
S.~M. Lundberg, S.-I. Lee, A unified approach to interpreting model
  predictions, in: I.~Guyon, U.~V. Luxburg, S.~Bengio, H.~Wallach, R.~Fergus,
  S.~Vishwanathan, R.~Garnett (Eds.), Advances in Neural Information Processing
  Systems 30, Curran Associates, Inc., 2017, pp. 4765--4774 (2017).

\bibitem{Ehlers2017}
R.~Ehlers, Formal verification of piece-wise linear feed-forward neural
  networks, in: D.~D'Souza, K.~Narayan~Kumar (Eds.), Automated Technology for
  Verification and Analysis, Springer International Publishing, Cham, 2017
  (2017).

\bibitem{Gore2014}
R.~Gor{\'{e}}, K.~Olesen, J.~Thomson, Implementing tableau calculi using bdds:
  Bddtab system description, in: Automated Reasoning - 7th International Joint
  Conference, {IJCAR} 2014, Held as Part of the Vienna Summer of Logic, {VSL}
  2014, Vienna, Austria, July 19-22, 2014. Proceedings, 2014, pp. 337--343
  (2014).

\bibitem{Cimatti2002}
A.~Cimatti, E.~Clarke, E.~Giunchiglia, F.~Giunchiglia, M.~Pistore, M.~Roveri,
  R.~Sebastiani, A.~Tacchella, {NuSMV Version 2: An OpenSource Tool for
  Symbolic Model Checking}, in: Proc. International Conference on
  Computer-Aided Verification (CAV 2002), Vol. 2404 of LNCS, Springer,
  Copenhagen, Denmark, 2002 (July 2002).

\bibitem{Caruana2015}
R.~Caruana, Y.~Lou, J.~Gehrke, P.~Koch, M.~Sturm, N.~Elhadad,
  \href{http://doi.acm.org/10.1145/2783258.2788613}{Intelligible models for
  healthcare: Predicting pneumonia risk and hospital 30-day readmission}, in:
  Proceedings of the 21th ACM SIGKDD International Conference on Knowledge
  Discovery and Data Mining, KDD '15, ACM, New York, NY, USA, 2015, pp.
  1721--1730 (2015).
\newblock \href {https://doi.org/10.1145/2783258.2788613}
  {\path{doi:10.1145/2783258.2788613}}.
\newline\urlprefix\url{http://doi.acm.org/10.1145/2783258.2788613}

\bibitem{Tornblom2019}
J.~Törnblom, S.~Nadjm-Tehrani, Formal Verification of Random Forests in
  Safety-Critical Applications: 6th International Workshop, FTSCS 2018, Gold
  Coast, Australia, November 16, 2018, Revised Selected Papers, 2019, pp.
  55--71 (01 2019).

\bibitem{Quinlan1993}
J.~R. Quinlan, C4.5: Programs for Machine Learning, Morgan Kaufmann Publishers
  Inc., San Francisco, CA, USA, 1993 (1993).

\bibitem{shannon1948mathematical}
C.~E. Shannon, A mathematical theory of communication, Bell system technical
  journal 27~(3) (1948) 379--423 (1948).

\bibitem{Cui2015}
Z.~Cui, W.~Chen, Y.~He, Y.~Chen,
  \href{http://doi.acm.org/10.1145/2783258.2783281}{Optimal action extraction
  for random forests and boosted trees}, in: Proceedings of the 21th ACM SIGKDD
  International Conference on Knowledge Discovery and Data Mining, KDD '15,
  ACM, New York, NY, USA, 2015, pp. 179--188 (2015).
\newblock \href {https://doi.org/10.1145/2783258.2783281}
  {\path{doi:10.1145/2783258.2783281}}.
\newline\urlprefix\url{http://doi.acm.org/10.1145/2783258.2783281}

\bibitem{Breiman2001}
L.~Breiman, Random forests, Machine Learning 45~(1) (2001) 5--32 (Oct 2001).

\bibitem{Freund1999}
Y.~Freund, R.~E~Schapire, A short introduction to boosting, Journal of Japanese
  Society for Artificial Intelligence 14 (1999) 771--780 (10 1999).

\bibitem{louppe2014}
G.~Louppe, Understanding random forests: From theory to practice, arXiv
  preprint arXiv:1407.7502 (2014).

\bibitem{de2011satisfiability}
L.~De~Moura, N.~Bj{\o}rner, Satisfiability modulo theories: introduction and
  applications, Communications of the ACM 54~(9) (2011) 69--77 (2011).

\bibitem{deMoura2008}
L.~de~Moura, N.~Bj{\o}rner, Z3: An efficient smt solver, in: C.~R.
  Ramakrishnan, J.~Rehof (Eds.), Tools and Algorithms for the Construction and
  Analysis of Systems, Springer Berlin Heidelberg, Berlin, Heidelberg, 2008,
  pp. 337--340 (2008).

\bibitem{Altmann2010}
A.~Altmann, L.~Toloşi, O.~Sander, T.~Lengauer,
  \href{https://doi.org/10.1093/bioinformatics/btq134}{{Permutation importance:
  a corrected feature importance measure}}, Bioinformatics 26~(10) (2010)
  1340--1347 (04 2010).
\newblock \href {https://doi.org/10.1093/bioinformatics/btq134}
  {\path{doi:10.1093/bioinformatics/btq134}}.
\newline\urlprefix\url{https://doi.org/10.1093/bioinformatics/btq134}

\bibitem{diabetes}
D.~Dua, C.~Graff, \href{http://archive.ics.uci.edu/ml}{{UCI} machine learning
  repository} (2017).
\newline\urlprefix\url{http://archive.ics.uci.edu/ml}

\bibitem{kick}
OpenML, Kick dataset, \url{https://www.openml.org/d/41162} (2019).

\bibitem{creditcard}
OpenML, Creditcard dataset, \url{https://www.openml.org/d/1597} (2019).

\bibitem{flight}
S.~Pafka, Flight dataset,
  \url{https://github.com/szilard/benchm-ml/tree/master/z-other-tools} (2019).

\bibitem{higgs}
P.~Baldi, P.~Sadowski, D.~Whiteson, Searching for exotic particles in
  high-energy physics with deep learning, Nature communications 5 (2014) 4308
  (07 2014).

\bibitem{silasparameters}
D.~I.~P. Ltd, Silas edu documentation,
  \url{https://www.depintel.com/documentation/_build/html/usage/learning/prep.html#parameters-in-settings}
  (2019).

\bibitem{h2oparameters}
H2O.ai, H2o dcoumentation: Build a random forest model,
  \url{http://docs.h2o.ai/h2o/latest-stable/h2o-r/docs/reference/h2o.randomForest.html}
  (2019).

\bibitem{rangerparameters}
M.~Wright, R dcoumentation: Ranger,
  \url{https://www.rdocumentation.org/packages/ranger/versions/0.11.2/topics/ranger}
  (2019).

\bibitem{heergithub}
N.~Heer, Speed comparison of programming languages,
  \url{https://github.com/niklas-heer/speed-comparison} (2019).

\bibitem{Abrahams2004}
D.~Abrahams, A.~Gurtovoy, C++ Template Metaprogramming: Concepts, Tools, and
  Techniques from Boost and Beyond (C++ in Depth Series), Addison-Wesley
  Professional, 2004 (2004).

\bibitem{Carmack2012}
J.~Carmack, In-depth: Functional programming in c++,
  \url{https://www.gamasutra.com/view/news/169296/Indepth_Functional_programming_in_C.php}
  (2012).

\end{thebibliography}


\end{document}